\documentclass[letterpaper, 10 pt, conference]{ieeeconf}  % Comment this line out if you need a4paper

\IEEEoverridecommandlockouts                              % This command is only needed if 
\usepackage{graphicx}
\usepackage{amssymb}
\usepackage{amsmath}
\usepackage{bm}
\usepackage{xspace}

\usepackage{url}
\usepackage{pgfplots}
\usepackage{tabularx}
\usepackage{multirow}
\usepackage{multicol}
\usepackage{threeparttable}
\usepackage{placeins}
\usepackage{balance}

\makeatletter
\let\NAT@parse\undefined
\makeatother
\usepackage[hidelinks]{hyperref}

\newcommand{\etal}{\xspace{}et al.\xspace}

\newcommand{\reffig}[1]{Fig.~\ref{#1}}
\newcommand{\refeq}[1]{Eq.~\ref{#1}}

\newcommand{\refsec}[1]{Section~\ref{#1}}

\title{\LARGE \bf
Real-Robot Deep Reinforcement Learning: Improving Trajectory Tracking of Flexible-Joint Manipulator with Reference Correction
}

\author{
	Dmytro Pavlichenko and Sven Behnke% <-this % stops a space
	\thanks{This work was funded by grant BE 2556/16-2 (Research Unit FOR 2535 Anticipating Human Behavior) of the German Research Foundation (DFG).}% <-this % stops a space
	\thanks{Both authors are with the Autonomous Intelligent Systems (AIS)
	Group, Computer Science Institute VI, University of Bonn, Germany. {\tt pavlichenko@ais.uni-bonn.de}}%
}

\usepackage{eso-pic}

\AtBeginDocument{\AddToShipoutPictureFG*{\AtTextUpperLeft{\put(0,\LenToUnit{10pt}){\parbox{\textwidth}{\centering\sffamily IEEE International Conference on Robotics and Automation (ICRA), Philadelphia, USA, May 2022.
}}}}}

\begin{document}
	
\maketitle
\thispagestyle{empty}
\pagestyle{empty}

\begin{abstract}
Flexible-joint manipulators are governed by complex nonlinear dynamics, defining a challenging control problem. In this work, we propose an approach to learn an outer-loop joint trajectory tracking controller with deep reinforcement learning. The controller represented by a stochastic policy is learned in under two hours directly on the real robot. This is achieved through bounded reference correction actions and use of a model-free off-policy learning method. In addition, an informed policy initialization is proposed, where the agent is pre-trained in a learned simulation. We test our approach on the 7\,DOF manipulator of a Baxter robot. We demonstrate that the proposed method is capable of consistent learning across multiple runs when applied directly on the real robot. Our method yields a policy which significantly improves the trajectory tracking accuracy in comparison to the vendor-provided controller, generalizing to an unseen payload.
\end{abstract}

% Contents
\section{Introduction}
\label{sec:Introduction}

The fundamental task for a robotic manipulator is to accurately follow the commanded trajectory, which is achieved by trajectory tracking control methods~\cite{arimoto_1990}~\cite{hou_2013}. The application domains of robotic manipulators are constantly expanding and new challenges arise. In particular, flexible-joint manipulators are used to ensure an increased safety for human workers in shared workspaces. It is not trivial to accurately control a flexible-joint manipulator, however, due to the complex nonlinear dynamics~\cite{sun_2019}. Design of classical controllers for such systems is time-consuming and often requires extensive instance and/or task-specific tuning.

\begin{figure}[t]
	\centering
	\includegraphics[width=8cm]{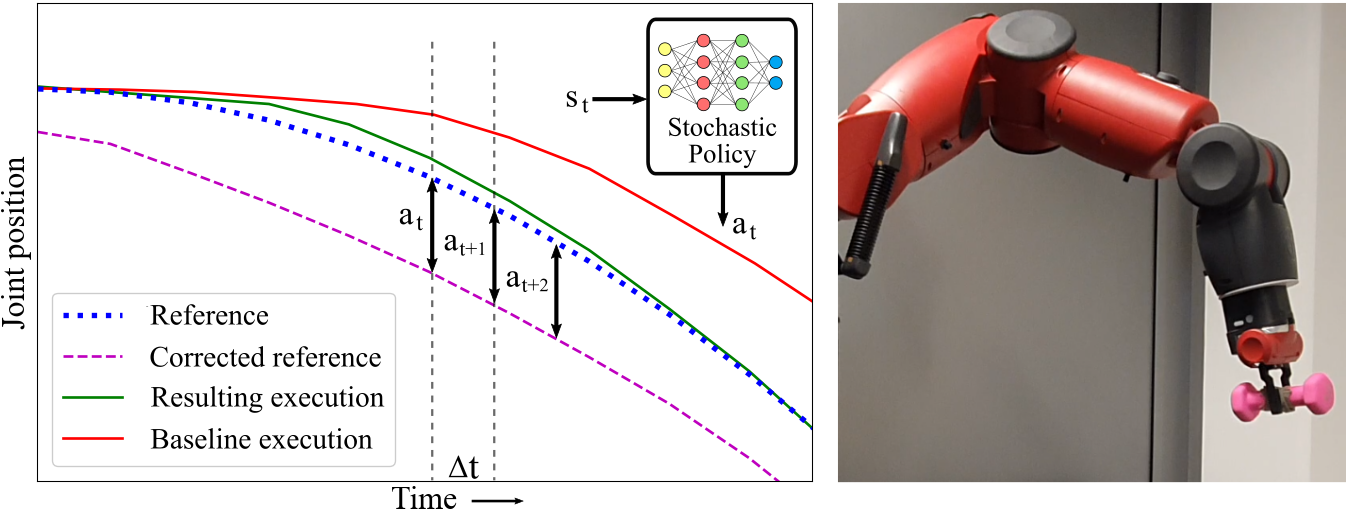}
	\caption{Stochastic reference correction policy learned directly on the real robot. Left: The policy learns to produce actions $a_t$ from the \mbox{states $s_t$}, which correct an arbitrary reference trajectory (blue) with \mbox{frequency $\Delta t$}, resulting in a corrected reference (magenta), which is fed to the vendor-provided controller and leads to an improved tracking accuracy (green) as opposed to the sole vendor-provided controller (red). Right: the policy is capable of compensating for an unseen payload.}
	\label{fig:teaser}
	\vspace*{-2ex}
\end{figure}

Deep reinforcement learning (DRL) methods produced effective policies for a broad range of control tasks~\cite{hwangbo_2019, rodriguez_2021, li_2021}. Most DRL approaches rely on learning in simulation and sim-to-real transfer. However, an accurate simulation of the robot is often not available. In this work, we propose an approach capable of learning an outer-loop control policy with DRL online, directly on the real robot. The policy operates at a lower frequency than the underlying classical controller and provides bounded reference correction actions. These corrections are applied to the reference trajectory before it is fed to the classical controller (\reffig{fig:teaser}). This formulation keeps the method agnostic of the underlying classical controller type. Our approach can also be interpreted as an online closed-loop trajectory optimization. The corrective actions of bounded magnitude alleviate safety concerns while training the model online on the real robot. To shorten the real-robot training time, we use an off-policy Soft Actor-Critic (SAC)~\cite{haarnoja_2018a} method. In order to further increase the learning speed, we propose an {\it informed initialization}: Policy pretraining in a {\it learned} simulation.

The evaluation is done on the 7 degrees of freedom (DOF) arm of a Baxter robot. The policy is learned in less than two hours. Our experiments demonstrate that addition of the learned high-level control policy significantly improves trajectory tracking accuracy in comparison to the vendor-provided controller. We also test how the learned policy handles a previously unseen change of the dynamics (attaching a payload) and the results indicate a persistent improvement of the trajectory tracking accuracy.

The key contributions of this work are:
\begin{itemize}
	\item Action, state and reward formulation to learn a reference correction policy directly on the real robot with DRL,
	\item informed initialization of the policy through a coarse dynamics
	model {\it learned} from data, which is used as a simulator.
\end{itemize}

\section{Related Work}
\label{sec:Related_Work}

Trajectory tracking control approaches can be divided into model-based~\cite{chae_1988} and model-free~\cite{longman_2000}. Model-free learning approaches are the popular tool to overcome the problem of imperfect models. The classical examples are Iterative Learning Control (ILC)~\cite{bristow_2006} and Repetitive Control (RC)~\cite{cuiyan_2004}. However, these methods are oriented on the repetitive execution of the exact same trajectories. Policies represented by Deep Neural Networks (DNN) obtained with Deep RL (DRL) were shown to generalize well to novel trajectories while approximating complex nonlinear dynamics~\cite{haarnoja_2018}.

First, we briefly review the most popular DRL algorithms. On-policy methods, such as: Trust Region Policy Optimization (TRPO)~\cite{schulman_2015}, Proximal Policy Optimization (PPO)~\cite{schulman_2017} and Asynchronous Advantage Actor-Critic A3C~\cite{mhin_2016} do not make use of past experience. Hence, a high sample complexity makes it challenging to perform learning directly on a real robot. In contrast, off-policy algorithms learn from past experience, drastically reducing sample complexity. Deep Deterministic Policy Gradient (DDPG)~\cite{lillicrap_2015} is a well-known example. It is, however, very challenging to tune. Soft Actor-Critic (SAC)~\cite{haarnoja_2018} alters the RL objective with a maximum entropy term~\cite{haarnoja_2017b} and is one of the newest off-policy algorithms, which we utilize in this work due to its sample efficiency and learning stability. 

There are numerous RL-based approaches to robotic manipulator control. Pradhan \etal~\cite{pradhan_2012} use actor-critic RL to adjust the model parameters in response to the payload variations on a two-link manipulator. Xu \etal~\cite{xu_2021} propose to learn a policy with DDPG to control a two-link manipulator in simulation. An RL agent acting as a nonlinear input compensator over the traditional controller is presented in~\cite{babuska_2014}. It is applied to a 1\,DOF robot in simulation. This idea is further explored by authors~\cite{babuska_2019}, being applied to a 5\,DOF UR5 manipulator, introducing a notion of RL-based reference compensation. A corrective controller is trained for each joint. Learning takes 30-70 executions for each trajectory, lacking generalization. Our action definition resembles the formulation of reference compensation. However, we train a single policy for all joints at once, which is advantageous when dealing with coupled flexible joints. Finally, we train the policy with DRL online on the real robot, promoting generalization to a broad range of trajectories. In some works, the approaches operate in task-space~\cite{yuke_2018}~\cite{lin_2019}. While this may be beneficial in certain cases, we focus on the joint space since it avoids redundancy.

On-policy RL methods such as PPO are used to train policies in simulation~\cite{hu_2018}~\cite{iriondo_2019}~\cite{kumar_2021}, which are then transferred to a real robot. This makes availability of an accurate simulation necessary. The resulting controller is used as a final product, which discards advantages of learning from the interaction with the real system. Moreover, in this case a re-traning is needed if the dynamics changes, for instance, due to wear and tear over time. Such re-training requires the aforementioned changes to be transferred back to the simulation first. In contrast, we present an approach to learn a policy directly on the real robot. This not only avoids the need for simulation, but allows to continuously learn from further experiences, adjusting to changes such as wear and tear. Finally, we augment the classical control law with a reference correction policy, instead of replacing it, creating an opportunity for both methods to complement each other.
\section{Background}
\label{sec:Background}

The objective of reinforcement learning is to find a policy $\pi$ which maximizes the expected discounted sum of rewards. In DRL, a policy $\pi_{\theta}$ is represented by a deep neural network, parameterized by learnable weights $\theta$. The problem is modeled as a Markov Decision Process (MDP): $\{S,A,P,r\}$ with state space $S \in \mathbb{R}^{n}$, action space $A \in \mathbb{R}^{m}$, state transition function $P \colon S \times A \mapsto S$ and a reward function $r \colon S \times A \mapsto \mathbb{R}$.

We address the problem with continuous state and action space, thus, we define a stochastic policy $\pi_{\theta}(\bm{a}|\bm{s})$, representing an action probability distribution when observing a state $\bm{s}_t$ at timesptep $t$. In this work, we use the off-policy SAC algorithm~\cite{haarnoja_2018} to train the policy. SAC is based on an actor-critic approach, where an actor provides actions and a critic represents the value function. SAC follows a maximum entropy RL formulation, which optimizes both expected reward and the entropy of the policy:
\begin{equation}
J(\pi_{\theta}) = \sum_{t=0}^{T} \mathbb{E} [r(\bm{s}_t, \bm{a}_t) + \alpha \: \mathcal{H}(\pi_{\theta}(\cdot | \bm{s}_t))],
\end{equation}
where $\mathcal{H}$ is the entropy of the stochastic policy and $\alpha \in [0, 1]$ is the temperature parameter. Note that with $\alpha = 0$ the above equation reduces to the conventional RL objective. The maximum entropy formulation incentivizes the policy exploration and was shown to produce robust polices and achieve stable learning. In this work, we use a SAC version with automatic tuning of the hyperparameter $\alpha$. The critic is realized with two Q-networks for increased stability of the learning process. A detailed description of the algorithm can be found in the original paper by Haarnoja \etal~\cite{haarnoja_2018}.
\section{Method}
\label{sec:Method}

\begin{figure}[t]
	\centering
	\includegraphics[width=8.4cm]{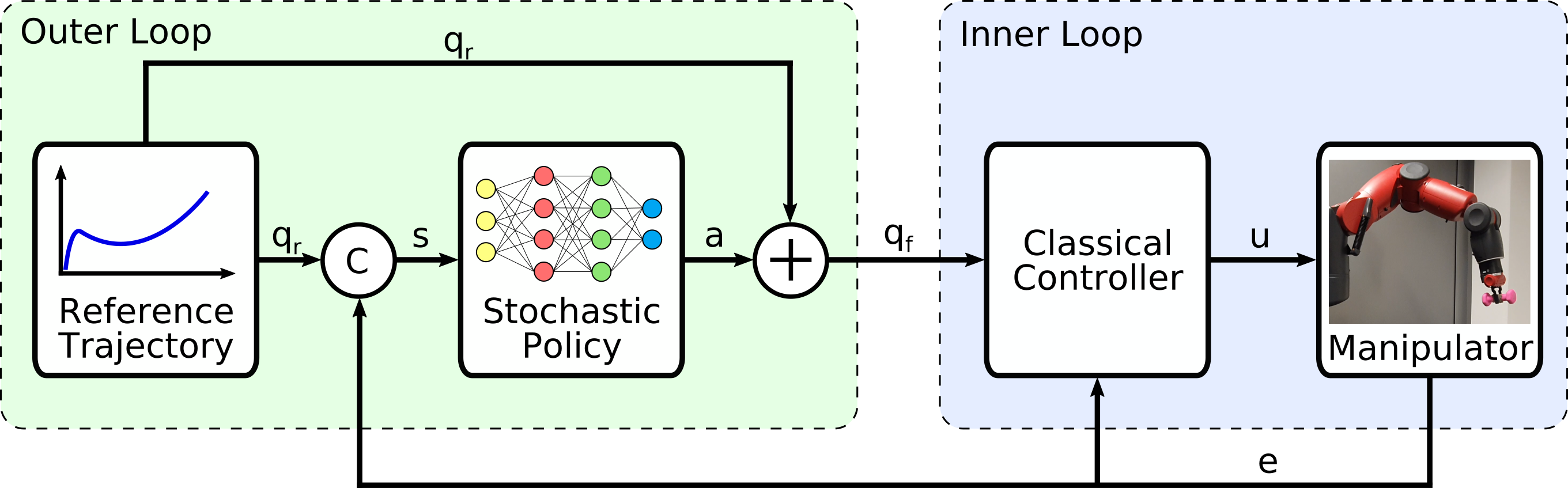}
	\caption{Control architecture. Given reference trajectory $q_r$ and feedback $e$, which form a state $s$, the learned stochastic policy corrects the reference $q_r$  with action $a$, resulting in a corrected reference $q_f$. The classical controller produces the control signal $u$ for the actuators. (C): concatenation.}
	\label{fig:control_diagram}
	\vspace*{-2ex}
\end{figure}

We propose an approach to learn a reference correction outer-loop controller to improve joint trajectory tacking accuracy over the existing classical controller (baseline) by utilizing DRL directly on a real robot. Given an arbitrary reference trajectory of joint positions $\bm{q}_r(t_0) \ldots \bm{q}_r(t_N)$ with $N$ equally spaced in time points with $\Delta t = t_i - t_{i-1}$ and duration $T = t_N - t_0$, our goal is to minimize the trajectory tracking error. Given a baseline with a subpar tracking accuracy, we propose to improve its performance by augmenting the control loop with an outer-loop learned policy. Underperformance of the baseline occurs when the available model is not accurate enough, the baseline was not tuned well enough, or the robot hardware accumulated significant wear and tear. Thus, achieving higher accuracy with the baseline requires to develop a better baseline or to perform tedious, instance and/or task-specific tuning. These solutions are time-consuming and require a highly skilled professional. We propose to utilize recent developments of DRL in order to learn a policy which compensates for the inaccuracies of the baseline (\reffig{fig:control_diagram}). The learning is done online on the real robot, which avoids the need for an accurate simulation and consequent sim-to-real transfer.

In the absence of an accurate model, DRL allows for efficient learning of the control policy, guided by the reward function. The trajectory tracking problem has {\it dense rewards}, which facilitates the learning process. Since we aim to perform the learning directly on the real robot, the DRL system should have high {\it robot persistence}~\cite{ibarz_2021}, the first component of which is {\it self-persistence}: the robot must not damage itself while training. We use strictly bounded actions which represent a corrective term for the reference trajectory. Thus, at any given time, the potential deviation from the reference trajectory is bounded. Selection of the maximum allowed magnitude of the corrective action provides a necessary flexibility when setting up the learning system. In addition, the corrective actions are filtered with a low-pass filter. Note, that we assume collision-free reference trajectories, hence the safety concerns raised here are related solely to the influence of models' actions. The second component is {\it task-persistence}: the robot must learn and collect data with minimal human assistance. The state of the manipulator at any time step is determined from the actuator encoders. Although there is noise, its magnitude is negligible since the learned policy operates at a relatively low frequency. Thus, there is no need for any additional software or hardware, which makes the learning setup and process coherent: the manipulator learns online from its own trial-and-error.

We chose a reference correction instead of input correction, because: First, input correction has to be done with a much higher frequency, which decreases the influence of a single action, obstructing the learning. Moreover, higher frequencies mean higher impact of latencies and noise, which makes the learning even more challenging. Second, reference correction is agnostic of the underlying baseline control law, which makes the approach more flexible.

\subsection{Action Space}
\label{sec:Action_Space}

The action $\bm{a}(t) \in \mathbb{R}^{N}$ for a robot with $N$ joints, produced by a learned stochastic continuous control policy, represents a reference compensation term: $\bm{q}_f(t+1) = \bm{q}_r(t+1) + \bm{a}(t)$, where $\bm{q}_r(t+1)$ is a reference trajectory point in joint space and $\bm{q}_f(t+1)$ is a resulting corrected reference. The corrective action $\bm{a}(t)$ produced at timestep $t$ is designated to change the next reference trajectory point at timestep $t+1$ such, that commanding the inner-loop classical controller to reach $\bm{q}_f(t+1)$ would result in the manipulator reaching $\bm{q}_r(t+1)$. We strictly bound the actions $\bm{a}(t) \in [-\bm{a}_{max}, \bm{a}_{max}]$ by a predefined constant vector $\bm{a}_{max}$. The stochastic policy runs at a frequency of 20\,Hz, so $\Delta t = t_i - t_{i-1} = 0.05$\,s. The produced actions $\bm{a}$ are filtered by a low-pass filter with a cutoff frequency of 4\,Hz to smooth out any inconsistent signal. The predefined constant $\bm{a}_{max}$ regulates the amount of power given to the learned policy. In this work, we choose to set $\bm{a}_{max} = \frac{0.95 \cdot \dot{\bm{q}}_{lim} \Delta t}{2}$, where $\dot{\bm{q}}_{lim}$ is a joint velocity limit vector. Thus, having two consecutive actions $\bm{a}(t) = -\bm{a}_{max}$ and $\bm{a}(t+1) = \bm{a}_{max}$ would satisfy the velocity limit $\dot{\bm{q}}_{lim}$. Indeed, as the corrective action is applied to the reference trajectory, this does not guarantee preserved velocity limits. However, it is a meaningful standardized formulation of $\bm{a}_{max}$ which can be used as a baseline value. Each point $\bm{q}_f(t)$ is checked for joint, velocity and acceleration limits (and clipped to them upon violation) before being commanded to the underlying controller.

\subsection{State Space}
\label{sec:State_Space}

The state $\bm{s}(t)$ of a robotic manipulator is a column vector:
\begin{equation}
\bm{s}(t) = {
	\begin{bmatrix}
	\bm{p}_o(t-2), \: \bm{a}(t-2), \: \bm{p}_o(t-1), \: \bm{a}(t-1),\\
	\bm{p}_o(t), \Delta \bm{p}_o(t), \\
	\bm{p}_r(t+1), \bm{p}_{r}(t+2)
	\end{bmatrix}
},
\label{eq:state}
\end{equation}
where point $\bm{p}$ contains joint positions and velocities: \mbox{$\bm{p} = [\bm{q}, \dot{\bm{q}}]$}, $\bm{p}_o$ is an observed point, read from the joint encoders, $\bm{p}_r$ is a point from the reference trajectory, \mbox{$\Delta \bm{p}_o(t) = \bm{p}_o(t) - \bm{p}_r(t)$} is an observed error. The first four components contain past observations and actions; the next two components contain current observations, augmented with the observed error; the last two components contain the future desired joint positions and velocities. Our state representation provides the policy with information about the past, the current state, and the future targets. Inclusion of the past observations and actions helps to combat the negative effects of latency~\cite{ibarz_2021}~\cite{riedmiller_2012}. In addition, it provides information about the dynamics of the manipulator, further strengthened by inclusion of joint velocities $\dot{\bm{q}}$. Several future reference points $\bm{p}_r$ provide more information about the desired motion of the manipulator. Each term in the state $\bm{s}(t)$ is rescaled to the $[-1, 1]$ interval. It is straightforward to do so for $\bm{q}$, $\dot{\bm{q}}$ and $\bm{a}$, given joint limits $\bm{q}_{lim}$, joint velocity limits $\dot{\bm{q}}_{lim}$, and action magnitude $\bm{a}_{max}$. There is no obvious way to choose an interpolation interval for $\Delta \bm{q}$ and $\Delta \dot{\bm{q}}$. In this work, we interpolate them with $[-(\mu_j + 3\sigma_j), \mu_j + 3\sigma_j]$, where $\mu_j$ and $\sigma_j$ are measured per-joint errors of positions and velocities, while executing random trajectories with the baseline controller.

\subsection{Reward Function}
\label{sec:Reward_Function}

For each timestep $t$, we define the reward function:
\begin{equation}
\label{eq:reward}
r(t) = \omega r_q(t) + (1 - \omega) r_v(t),
\end{equation}
where $r_q \in [0, 1]$ is a reward term encouraging joint position tacking, $r_v \in [0, 1]$ is a reward term encouraging joint velocity tracking, and $\omega \in [0, 1]$ is a relative importance scaling factor. Given that, $r \in [0, 1]$. To compute $r_q$ and $r_v$, we first define the cumulative absolute errors $e_q$ for the joint position and $e_v$ for the joint velocity:
\begin{equation}
\label{eq:error}
e_q(t) = \sum_{j=0}^{j=N} |\bm{q}_o^j(t) - \bm{q}_r^j(t)|,
\end{equation}
where $N$ is the number of joints, $\bm{q}_o^j(t)$ is the observed position of joint $j$, and $\bm{q}_r^j(t)$ is the desired position at time step $t$. The procedure is analogous for $e_v$. Finally, we use the smooth logistic kernel function $K$~\cite{rodriguez_2021} to define the reward terms:
\begin{equation}
	K(x, l)=\frac{2}{e^{xl} + e^{-xl}},
\end{equation}
\begin{equation}
r_q(t) = K(e_q(t), l_q),
\end{equation}
where $l$ is the kernel sensitivity parameter. The term $r_v(t)$ is defined analogously. When a trajectory is tracked perfectly, $r=1$. We use the L1 norm in \refeq{eq:error}, because when using the L2 norm we noticed that while learning, the policy would aggressively abuse joints with smaller errors to compensate for joints with bigger errors, leading to learning instabilities and unintuitive motions of the manipulator.

We include the joint velocity tracking term $r_v(t)$ into the reward function to promote trajectory smoothness. We observed that without this term, the policy would often learn a bang-bang style control, leading to non-smooth motions. Adding an explicit action cost term was extremely hard to tune, and often led to an abusive behavior, when the policy would generally avoid making any corrections with sudden corrections of large magnitude in between. Finally, we concluded that rewarding joint velocity tracking is a natural way to promote trajectory smoothness though following the derivative of the reference path. Since our main goal is to improve joint position tracking accuracy, we set $\omega = 0.75$.

\subsection{Model}
\label{sec:Model}

In this work, we train the stochastic policy using the SAC algorithm~\cite{haarnoja_2018}. The method is off-policy, making this choice natural for the real-robot learning due to its increased sample efficiency. Unlike the original SAC, where actions are bounded to a finite range through the use of a Gaussian policy with the squashing function, we use a Beta policy~\cite{wei_2017}, which is bounded in the $[0, 1]$ range by definition. Both actor and critic networks have similar structure: they consist of two hidden fully-connected layers with tanh nonlinearity. The output layer of the Q-networks provides a single Q-value, when supplied with a state-action pair $\{\bm{s}(t), \bm{a}(t)\}$. The actor network outputs two values for each dimension of the action space, resulting in $2 \cdot |\bm{a}|$ outputs. Each pair of these values parameterizes a beta distribution, from which the actions are sampled. The sampled actions are in the $[0, 1]$ range. It is straightforward to rescale them to $[-\bm{a}_{max}, \bm{a}_{max}]$. We use the sigmoid activation function in the output layer of the actor network. Since beta distribution parameters $\in (0, \infty]$, the outputs of the actor are clipped to $[\epsilon, 1]$, where $\epsilon = 1 \times 10^{-5}$. Finally, we scale them up by a factor of 10 to provide the actor with enough freedom for distribution selection. During training, the actions are sampled from the distributions, while during inference we use their modes.

\subsection{Learning Process}
\label{sec:Learning_Process}

A single episode corresponds to a reference trajectory consisting of $T$ joint positions $\bm{q}(t)$ and velocities $\dot{\bm{q}}(t)$, equally spaced in time with interval $\Delta t$. Thus, every episode is finite by definition. During each episode, at each time step $t$ with the observed state $\bm{s}_t$ the policy provides a reference correction action $\bm{a}(t)$, leading to state $\bm{s}(t+1)$ with reward $r(t)$. The tuple $\{\bm{s}(t), \bm{a}(t), r(t), \bm{s}(t+1)\}$ is stored in the replay buffer for experience replay. We perform SAC update iterations with a replay ratio~\cite{fedus_2020} of 1 (one update per one point added to the buffer). The weights of the actor network are updated after each episode. For each SAC iteration, a minibatch of uniformly sampled datapoints is generated.

\subsection{Informed Initialization}
\label{sec:Init_Sim}

It is beneficial to have a deliberate initialization of the model before applying it to the real robot, since a randomly initialized policy yields a low-reward behavior. In the absence of an accurate simulation, we propose to learn a coarse simulation, represented by a neural network to pretrain the policy. First, a small sample of random trajectories from the real robot is recorded. Then, we train a neural network consisting of three fully connected layers to output \mbox{$[\bm{q}_o(t+1), \dot{\bm{q}}_o(t+1)]$}, given \mbox{$[\bm{q}_o(t), \dot{\bm{q}}_o(t)]$} and \mbox{$[\bm{q}_r(t+1), \dot{\bm{q}}_r(t+1)]$}. Finally, we use this coarse forward dynamics model as a simulation environment to pretrain the stochastic policy before using it on the real robot. This pretraining step keeps the approach generic, as it does not require any additional knowledge about the manipulator dynamics, while representing latencies and noise from the real system, informing the policy about its future experiences. Thus, we refer to this approach as informed initialization. The training in this learned simulation and on the real robot is performed as described in \refsec{sec:Learning_Process}. In addition, the data gathered for informed initialization is used to determine scaling factors for $\Delta \bm{q}$ and $\Delta \dot{\bm{q}}$ (\refsec{sec:State_Space}).
\section{Evaluation}
\label{sec:Evaluation}

To evaluate the proposed approach, we apply it to the left arm of a Baxter robot. It is a 7\,DOF manipulator with flexible joints. The underlying dynamics model is unknown and flexible joints with coupled dynamics are challenging to control. We aim to answer the following questions:
\begin{itemize}
	\item Does our approach improve the joint position trajectory tracking accuracy?
	\item Does it generalize to handle an unseen payload?
	\item Is the learning consistent and safe for the hardware?
\end{itemize}

\subsection{Setup}

\begin{figure}[t!]
	\centering
	{\footnotesize a)}\includegraphics[width=4.03cm]{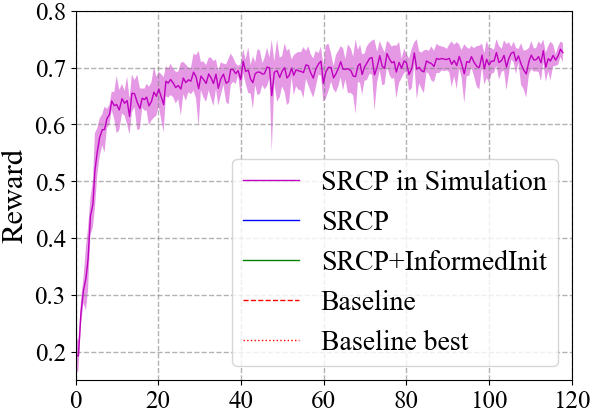}
	{\footnotesize b)}\includegraphics[width=4.03cm]{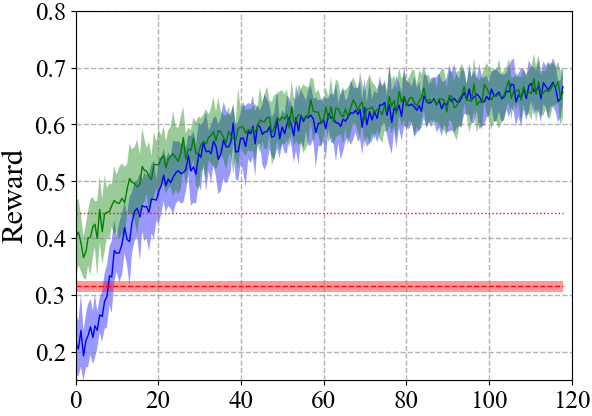}
	{\footnotesize c)}\includegraphics[width=4.03cm]{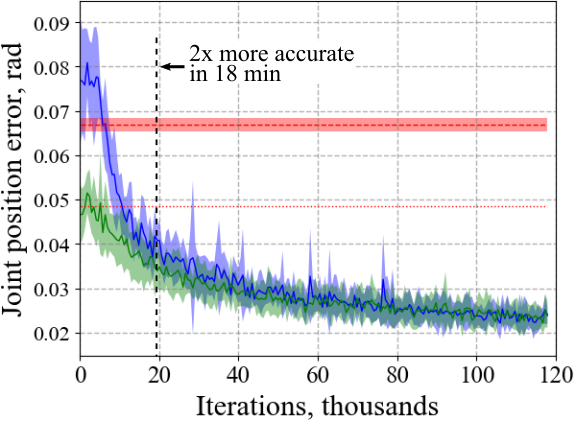}
	{\footnotesize d)}\includegraphics[width=4.03cm]{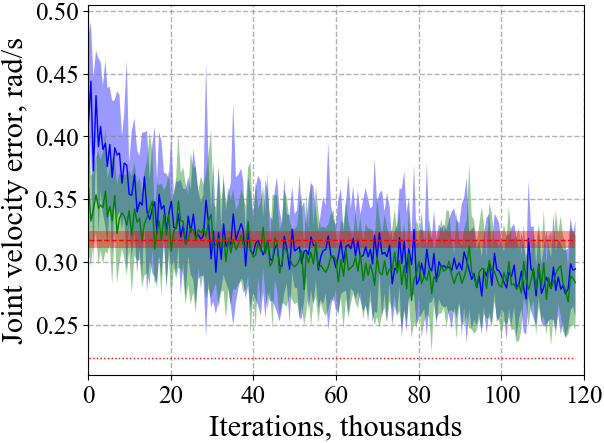}
	\caption{Training on 1000 trajectories. a) SRCP reward in learned simulation. b) Rewards on the real robot. c) Average joint position tracking error. d) Average joint velocity tracking error. Both variants of the approach achieve stable learning. Solid lines represent the mean and shaded regions represent 95\% confidence intervals, averaged over three runs. Baseline best: the best result achieved by the baseline controller. Average iterations per trajectory: 118. 10K iterations $\approx$ 9\,min of real time.}
	\label{fig:reward}
	\vspace*{-2ex}
\end{figure}

The classical inner-loop controller which we further refer to as baseline is the Baxter inverse dynamics controller\footnote{\url{https://sdk.rethinkrobotics.com/wiki/Joint_Trajectory_Action_Server}}, which calculates commanded torques from the supplied joint positions and velocities. We chose it as the baseline in our experiments because it showed the best joint position tracking accuracy, compared to the Baxter position and velocity PD controllers. In the case of this baseline we employ reference correction for both joint position and velocities, hence $\bm{a}(t) = [\bm{a}_q(t), \bm{a}_v(t)] \in \mathbb{R}^{2N}$, where $\bm{a}_q(t)$ is the reference joint position correction and $\bm{a}_v(t)$ is the reference joint velocity correction. With $N=7$ this results in a 14-element action vector. The maximum magnitude of the velocity corrective term $\bm{a}_{{max}_{v}}$ was defined as described in \refsec{sec:Action_Space}, only using joint acceleration limits.

According to our state representation (Eq.~\ref{eq:state}) and the provided 14-element $\bm{a}$ and $\bm{p}$, we obtain a state vector of \mbox{$8 \times 14 = 112$} elements. We use the same size of 80 neurons for hidden layers for both critic and actor networks. This results in a \mbox{$112 \times 80 \rightarrow 80 \times 80 \rightarrow 80 \times 28$} actor network and a \mbox{$(112+14) \times 80 \rightarrow 80 \times 80 \rightarrow 80 \times 1$} critic network. For training, we use Adam optimizer with triangular learning rate scheduling, ranging from $1 \times 10^{-4}$ to  $4 \times 10^{-4}$ with a period of 100 episodes. We set the discount factor $\gamma = 0.85$ and perform a hard critic update with $\tau = 1$ every 1000 iterations. We use a minibatch of 128 datapoints. The kernel sensitivities were set to $l_q = 32$ for the joint position tracking reward term and $l_v = 7$ for the joint velocity tracking reward term. Since the reward scale is critical for SAC due to the entropy maximization~\cite{haarnoja_2018}, we set its value to 10, as we empirically found that larger or smaller values resulted in inferior learning. The training is performed on a regular laptop with Intel i7-6700HQ CPU and 16 GB of RAM.

\subsection{Experiments}

To learn the Stochastic Reference Compensation Policy (SRCP), we generate random trajectories from the workspace of Baxter's left arm. We cover the region $Y$ in front and to the side of the robot with approximate dimensions of $1.4 \times 0.7 \times 1.0$\,m. Each trajectory is defined by a start and a goal plus 1-3 intermediate points. Each point represents a 6D end-effector pose and is drawn randomly from $Y$. An inverse kinematics solver is used to convert trajectories to joint space. A trajectory is checked for joint, velocity and acceleration limits, as well as for collisions before execution. If found to be unfeasible, a new trajectory is generated.

To evaluate the effectiveness of the informed initialization, as proposed in \refsec{sec:Init_Sim}, we also train the forward dynamics model to represent the simulation. The model has three fully connected layers: \mbox{$28 \times 64 \rightarrow 64 \times 32 \rightarrow 32 \times 14$}. We train this model with Adam optimizer and learning rate of $10^{-4}$ until convergence on a small real-robot dataset (20\,min duration) of 200 trajectories generated as described above. Finally, we train two variants of SRCP: one with random weights and the other one is first pretrained in the learned simulation. We refer to the latter as SRCP+InformedInit.

\begin{figure}[t!]
	\centering
	{\footnotesize a)}\includegraphics[width=4.03cm]{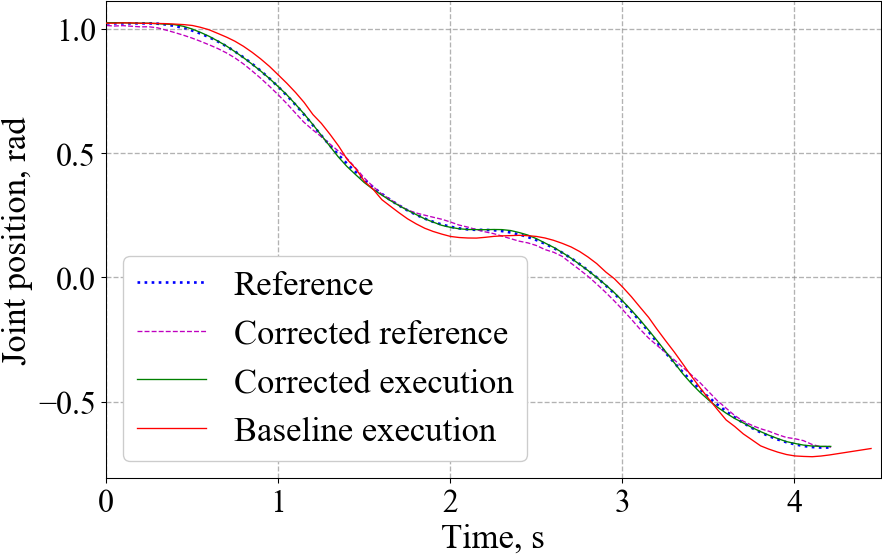}
	{\footnotesize b)}\includegraphics[width=4.03cm]{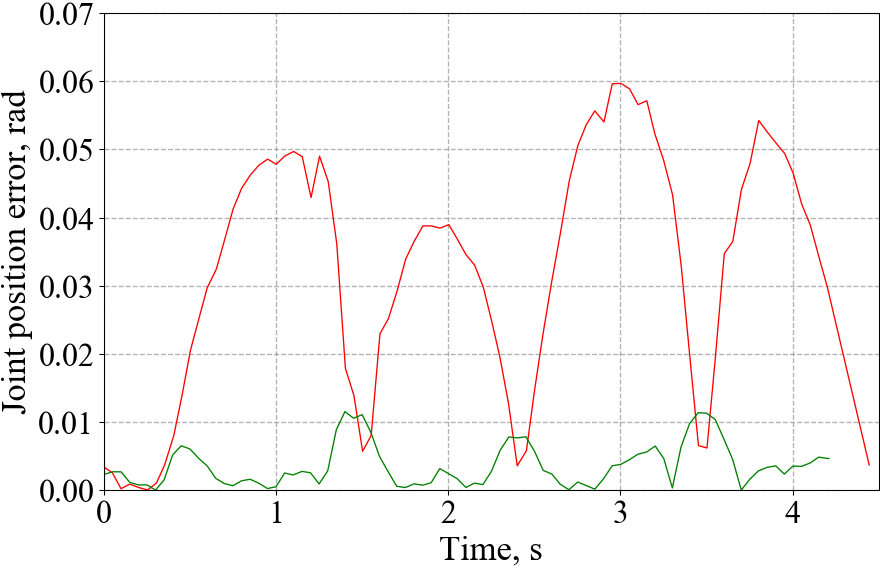}
	{\footnotesize c)}\includegraphics[width=4.03cm]{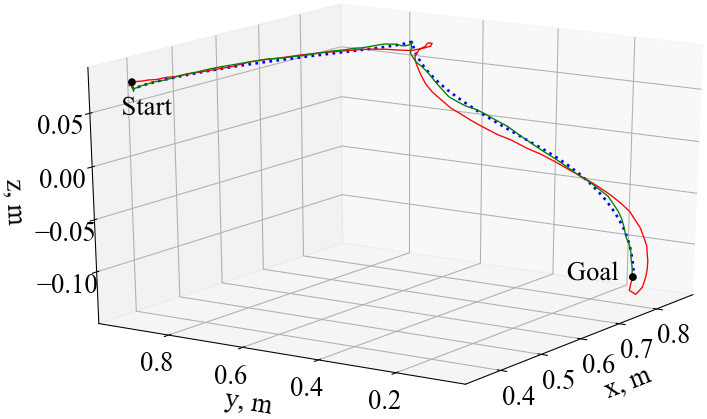}
	{\footnotesize d)}\includegraphics[width=4.03cm]{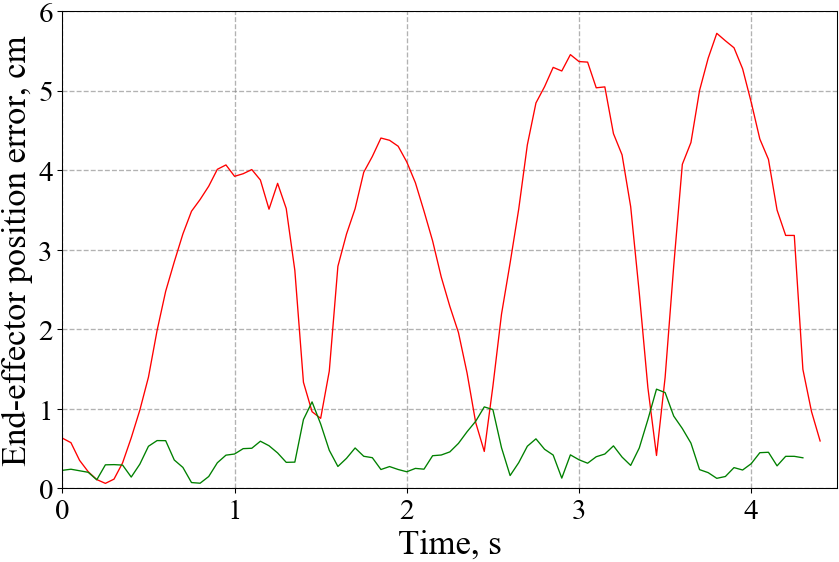}
	\caption{Example test trajectory. a) Shoulder yaw position trajectory. b) Tracking error of shoulder yaw. c) Path of the end-effector. d) End-effector position tracking error. While the baseline controller (red line) leads to significant deviations from the desired trajectory (blue dotted line), combination with our learned controller leads to a more accurate tracking (green line), achieved by reference compensation (magenta dashed line).}
	\label{fig:trajectories}
	\vspace*{-2ex}
\end{figure}

We perform three training runs for each model. A run consists of 1000 random trajectories. One run completes in under 2 hours, with approximately 100\,min dedicated for the trajectory execution. In \reffig{fig:reward}, we show the training reward curves as well as the joint position and joint velocity tracking errors, averaged over the three runs. Since each trajectory has a different number of points, the cumulative values would not be representative; and we show all values averaged over the number of points. The baseline statistics were calculated from 100 random trajectories. While training in learned simulation (takes less than 20\,min), the SRCP quickly reaches relatively high rewards (\reffig{fig:reward}a). On the real robot, the policy learns slower (\reffig{fig:reward}b) due to the latencies and complex dynamics. One can see that SRCP+InformedInit directly starts in the reward region above the baseline, achieving higher rewards faster than the randomly initialized policy. Note, that SRCP+InformedInit becomes twice more accurate than the baseline only in 18\,min. Both policies converge to a similar performance in the end. The average cumulative joint position error $e_q$ for a trajectory with $N$ joints consisting of $T$ points is calculated according to \refeq{eq:error}, additionally averaged by $T$. The average cumulative joint velocity error is calculated analogously. One can observe that the joint position tracking error, which we prioritize in \refeq{eq:reward}, is being reduced considerably during training. The joint velocity tracking error declines as well, serving as a smoothness constraint. SRCP inference took $0.005 \pm 0.002$\,s on average, not causing noticeable delay compared to the original control loop. The entropy temperature parameter converged to $\alpha = 0.05 \pm 0.01$.

\begin{table}[]
	\centering
	\caption{Average joint position tracking error, rad $\times 10^{-2}$.}
	\label{table:joint_error}
	\begin{tabular}{ccc|cc}
		\multicolumn{1}{l}{} & \multicolumn{2}{c|}{No payload}     & \multicolumn{2}{c}{0.9 kg payload}  \\ \hline
		Joint                & Baseline        &  B+SRCP           & Baseline         & B+SRCP           \\ \hline
		1                    & 2.75 $\pm$ 1.83 &  0.52 $\pm$ 0.43  & 3.32 $\pm$ 2.32  & 0.47 $\pm$ 0.42  \\
		2                    & 1.13 $\pm$ 1.08 &  0.36 $\pm$ 0.27  & 2.31 $\pm$ 1.73  & 1.17 $\pm$ 0.62  \\
		3                    & 1.07 $\pm$ 0.87 &  0.21 $\pm$ 0.19  & 1.38 $\pm$ 1.10  & 0.27 $\pm$ 0.25  \\
		4                    & 0.66 $\pm$ 0.71 &  0.19 $\pm$ 0.20  & 0.79 $\pm$ 0.74  & 0.25 $\pm$ 0.20  \\
		5                    & 0.51 $\pm$ 0.35 &  0.27 $\pm$ 0.22  & 0.62 $\pm$ 0.53  & 0.38 $\pm$ 0.32  \\
		6                    & 0.38 $\pm$ 0.28 &  0.31 $\pm$ 0.27  & 0.62 $\pm$ 0.52  & 0.41 $\pm$ 0.37  \\
		7                    & 0.35 $\pm$ 0.26 &  0.19 $\pm$ 0.15  & 0.30 $\pm$ 0.24  & 0.16 $\pm$ 0.14  \\ \hline
		$\sum$               & 6.87 $\pm$ 3.15 &  2.08 $\pm$ 0.87  & 9.36 $\pm$ 4.07  & 3.13 $\pm$ 1.16  \\
		%&                    & (70\%)          &                   & (67\%)  
	\end{tabular}
	\begin{tablenotes}
		\item \hspace{1ex} \scriptsize{*Mean $\pm$ SD is shown. B+SRCP stands for Baseline+SRCP.}
	\end{tablenotes}
	\vspace*{-3ex}
\end{table}

We evaluate the learned SRCP by executing 100 unseen test trajectories and comparing the joint position tracking accuracy of the \mbox{baseline + SRCP} control against the vendor-provided baseline. In addition, we perform the same test with a payload of 0.9\,kg to evaluate the performance in the presence of altered dynamics (maximum payload for Baxter is 2.2\,kg). A video is available online\footnote{\url{https://www.ais.uni-bonn.de/videos/ICRA_2022_Pavlichenko}}. We show the measured average joint position tracking error in Table~\ref{table:joint_error} and end-effector position tracking error in Table~\ref{table:cartesian_error}. Joints are labeled from 1 to 7, going from the shoulder to the wrist. SRCP improves the trajectory tracking accuracy more than three times for both cases: without and with previously unseen payload, achieving an average end-effector tracking error of 0.66\,cm without a payload (more than four times more accurate than the baseline). For comparison, advanced MPC controllers applied to Baxter~\cite{rupert_2015}~\cite{terry_2017} achieved 1-2.5\,cm average steady-state error. The proposed method outperformed our earlier work based on offline supervised learning~\cite{pavlichenko_2021}. There is a dependency between position of the joint in the kinematic chain and its tracking error: first joints have a much higher error. This effect can be explained by a larger inertia affecting the base joints. Moreover, one can observe that joint 2 has the highest increase in tracking error when the payload is added. We explain this by the fact that it has a passive external spring, which further complicates its dynamics. An example trajectory of the shoulder yaw position and the resulting end-effector path are shown in \reffig{fig:trajectories}. The baseline controller frequently deviates from the reference trajectory while accelerating (seconds 0-1 and 2.5-3.5) or decelerating (seconds 1-2 and 3.5-4.5). The addition of the learned SRCP compensates for these deviations.

Overall, the conducted experiments shown that the proposed approach consistently improved the policy directly on the real robot. The proposed informed initialization through a learned simulation significantly reduced the tracking error at the beginning of learning. Thanks to the bounded reference correction actions, filtered by a low-pass filter, we did not observe jerky motions while training. The resulting stochastic policy improved the trajectory tracking accuracy more than three times, compared to the baseline, achieving sub-centimeter accuracy of the end-effector. Finally, it demonstrated a persistent improvement of accuracy in the experiment with a previously unseen payload, suggesting that the policy learned to use live feedback from the robot, instead of simply memorizing the necessary corrections.

\begin{table}[]
	\centering
	\caption{Average end-effector position tracking error, cm.}
	\label{table:cartesian_error}
	\begin{tabular}{cc|cc}
		\multicolumn{2}{c|}{No payload}      & \multicolumn{2}{c}{0.9 kg payload} \\ \hline
		Baseline          & B+SRCP           & Baseline         & B+SRCP          \\ \hline
		3.12 $\pm$ 1.81   & 0.66 $\pm$ 0.42  & 4.35 $\pm$ 2.32  & 1.19 $\pm$ 0.60 \\  
		%& (79\%)          &                  & (73\%)  
	\end{tabular}
	\begin{tablenotes}
		\item \hspace{1ex} \scriptsize{*Mean $\pm$ SD is shown. B+SRCP stands for Baseline+SRCP.}
	\end{tablenotes}
	\vspace*{-3ex}
\end{table}

\subsection{Discussion}

We observed that the discount factor $\gamma$ significantly influences the learning process. For our case, we found $\gamma \in [0.75, 0.85]$ to be the best. A small value, such as \mbox{$\gamma = 0.5$} would typically result in a poor performance of the velocity component $r_v$ and non-smooth trajectories. High values, like $\gamma \in [0.95, 0.99]$ resulted in a much slower learning. We explain this observation by the fact that in our problem setting the policy does not have full control over the state: it is always tied to the reference. Since the policy does not have access to the complete reference trajectory at once, accounting for the rewards which are far in the future makes the learning more challenging due to the inherent uncertainty.

There are two task-specific aspects in the presented approach. First, the low-pass filter of the actions introduces an additional delay which the model has to learn. However, removing filtering led to frequently occurring jerky motions. Second, the scaling of $\Delta \bm{q}$ and $\Delta \dot{\bm{q}}$ components, which is determined empirically. In contrast to the filtering, these components can be removed for increased generality of the method without significantly harming the performance. However, we observed a speed-up of the learning process when including these components in the state representation.

The key hyperparameter for our approach is the maximum magnitude $\bm{a}_{max}$ of the corrective action $\bm{a}$. While it is not trivial to choose its value, it provides an increased control over the learned policy. For example, in cases when the method is tried for the first time on a new manipulator, it is easy to safely test the approach, using a small value of $\bm{a}_{max}$. One interesting direction for the future work is to have $\bm{a}_{max}$ learned together with the stochastic policy. Additionally, introducing adaptive constraints on the tracking error while learning, would improve the safety of the approach.
\section{Conclusion}	
\label{sec:Conclusion}

We presented a model-free approach to learn a stochastic policy for improving joint trajectory tracking accuracy of a flexible-joint manipulator. The learning is performed with deep reinforcement learning directly on the real robot. The obtained policy serves as an additional outer-loop controller and provides reference correction actions. In addition, we propose to perform an informed initialization which uses a learned coarse forward dynamics model as a simulation. We demonstrate that the proposed method is capable of consistent learning on the real 7\,DOF manipulator of a Baxter robot without the need for an accurate hand-crafted simulation and a consecutive sim-to-real transfer. Our experiments indicate that the policy learned in under two hours improves the trajectory tracking accuracy by more than a factor of three over the vendor-provided baseline controller. The learned policy is general enough to demonstrate a persistent improvement when dealing with an unseen payload. Finally, the proposed informed initialization made the policy learn high-reward behaviors faster than a random initialization.

% Bibliography
\bibliographystyle{IEEEtran}
\bibliography{bibliography}

\end{document}